%% file: GSIP_v16.tex
\documentclass{article}
\pdfoutput=1
\usepackage{amsmath}
\usepackage{amsfonts}
\usepackage{amssymb}
\usepackage{graphicx}
\usepackage{spconf}
\usepackage[T1]{fontenc}    
\usepackage{hyperref}       
\usepackage{url}            
\usepackage{booktabs}       
\usepackage{nicefrac}       
\usepackage{microtype}      
\usepackage{bbm}
\usepackage{amsthm}
\usepackage{cite}
\input{commands}
\usepackage{multirow}

\usepackage{etoolbox}

\makeatletter
\@tempswafalse
\@ifpackageloaded{amsthm}{\@tempswatrue}{}
\@ifpackageloaded{ntheorem}{\@tempswatrue}{}
\if@tempswa
  \patchcmd\@thm{\trivlist}{\small\trivlist}{}{}
\else
  \patchcmd\@begintheorem{\trivlist}{\small\trivlist}{}{}
  \patchcmd\@opargbegintheorem{\trivlist}{\small\trivlist}{}{}
\fi
\makeatother

\newtheorem{theorem}{Theorem}[]
\newtheorem{lemma}[theorem]{Lemma}
\usepackage{color}
\usepackage[font=small,labelfont=bf]{caption}

\newcommand{\startcompact}[1]{\par\vspace{-1em}\begin{#1}%
\allowdisplaybreaks\ignorespaces}

\newcommand{\stopcompact}[1]{\end{#1}\ignorespaces}

\theoremstyle{definition}
\newtheorem{definition}{\small Definition D.\ignorespaces}[]

\newtheorem{assumption}{\small Assumption A.\ignorespaces}[]

\newtheorem*{remark*}{Remark}

\newcommand{\jdh}[1]{}
\renewcommand{\jdh}[1]{{\color{red}{#1}}} 

\makeatletter
\g@addto@macro \small {%
 \setlength\abovedisplayskip{4pt plus 2pt minus 2pt}%
 \setlength\belowdisplayskip{4pt plus 2pt minus 2pt}%
}
\makeatother

\makeatletter
\g@addto@macro \footnotesize {%
 \setlength\abovedisplayskip{4pt plus 1pt minus 1pt}%
 \setlength\belowdisplayskip{4pt plus 1pt minus 1pt}%
}
\makeatother

\makeatletter
\g@addto@macro \normalsize {%
 \setlength\abovedisplayskip{5pt plus 3pt minus 3pt}%
 \setlength\belowdisplayskip{5pt plus 3pt minus 3pt}%
}
\makeatother

\newcommand{\edit}[1]{\textcolor{black}{#1}}

\usepackage{hyperref}
\title{A Dictionary based Generalization of Robust PCA \vspace{-0.7em}}
	\name{Sirisha Rambhatla, Xingguo Li and Jarvis Haupt\vspace{-1em } \thanks{The authors graciously acknowledge support from NSF Award CCF-1217751 and the DARPA Young Faculty Award, Grant N66001-14-1-4047. }}
\address{Department of Electrical and Computer Engineering, \\ University of Minnesota-Twin Cities, Minneapolis, MN-55455\\\vspace{-0.5em }  {\small\texttt{\{rambh002, lixx1661, jdhaupt\}@umn.edu}} \vspace{-0.7em }}

\begin{document}
\allowdisplaybreaks
\maketitle
\begin{abstract}
\vspace{-4pt}
We analyze the decomposition of a data matrix, assumed to be a superposition of a low-rank component and a component which is sparse in a known dictionary, using a convex demixing method. We provide a unified analysis, encompassing both undercomplete and overcomplete dictionary cases, and show that the constituent components can be successfully recovered under some relatively mild assumptions up to a certain \textit{global} sparsity level. Further, we corroborate our theoretical results by presenting empirical evaluations in terms of phase transitions in rank and sparsity for various dictionary sizes.
\end{abstract}
\vspace{-5pt}
\vspace{-10pt}
	\section{Introduction}
\vspace{-8pt}
	Exploiting the inherent structure of data for the recovery of relevant information is at the heart of data analysis. 
	%
	In this paper, we analyze a \edit{scenario where a data} matrix {\small$\mb{Y} \in \mathbb{R}^{n \times m}$} arises as a result of a superposition of a \edit{rank-{\small $r$}} component {\small$\mb{X} \in \mathbb{R}^{n \times m}$}, and a dictionary sparse component expressed here as {\small$\mb{RA}$}. Here, {\small$\mb{R} \in \mathbb{R}^{n \times d}$} is an \textit{a priori} known dictionary with normalized columns, and {\small$\mb{A} \in \mathbb{R}^{d \times m}$} is the \edit{unknown} \textit{sparse} coefficient matrix with \edit{at most} {\small $s$} total non-zeros. Specifically, we will study the following model,
	\startcompact{small}
	\begin{align}
	\label{Prob}
	\mb{Y} = \mb{X} + \mb{RA},
	\end{align}
	\stopcompact{small}%
	and identify the conditions under which the components {\small$\mb{X}$} and {\small$\mb{A}$}, can be successfully recovered, given {\small$\mb{Y}$} and {\small$\mb{R}$}. 	
	 
	 A wide range of problems can be expressed in the form described above. Perhaps the most celebrated of these is principal component analysis (PCA) \cite{Jolliffe02}, which can be viewed as a special case of eq.\eqref{Prob}, with the matrix {\small$\mb{A}$} set to zero. In the absence of the component {\small$\mb{X}$}, the problem reduces to that of sparse recovery \cite{Natarajan95, Donoho01, Candes05}; See \cite{Rauhut10} and references therein for an overview of related works. The popular framework of Robust PCA tackles a case when the dictionary {\small$\mb{R}$} is an identity matrix \cite{Candes11, Chandrasekaran11}; variants include \cite{Zhou2010, Ding11, Wright13, Chen13}. In addition, other variants of Robust PCA, such as Outlier Pursuit \cite{Xu2010}, where {\small$\mb{R} = \mb{I}$} and the sparse component is column sparse, and randomized adaptive sensing approaches \cite{Li15, Li15c, Li15b, Rahmani2015, Li16}, have also been explored.
	 
	 %
	 Our work is \edit{most} closely related to \cite{Mardani2012}, which explores the application of the model shown in eq.\eqref{Prob} to detect traffic anomalies, and focuses on a case where the dictionary {\small$\mb{R}$} is \textit{overcomplete}, i.e., \textit{fat}. The model described therein,  is applicable to a case where the rows of {\small$\mb{R}$} are orthogonal, e.g., {\small$\mb{RR'} = \mb{I}$}, and the coefficient matrix {\small$\mb{A}$}, has at most {\small $k$} nonzero elements per row and column. \edit{In this paper, we analyze a more general case, where we relax some of the aforementioned assumptions for the \textit{fat} case, and develop an analogous analysis for the \textit{thin} case. Specifically, this paper makes the following contributions towards guaranteeing the recovery of {\small$\mb{X}$} and {\small$\mb{A}$} in eq.\eqref{Prob}. First, we analyze the \textit{thin} case, where we assume {\small$\mb{R}$} to be a \textit{frame} \cite{Duffin1952} with a \textit{global} sparsity of at most {\small $s$}; See \cite{Heil2013} for a brief overview of frames. Next, for the \textit{fat} case, we extend the analysis presented in \cite{Mardani2012}, and assume that the dictionary {\small$\mb{R}$} satisfies the restricted isometry property (RIP) of order {\small $k$}, with a global sparsity of at most {\small $s$}, and a column sparsity of at most {\small $k$}. Consequently, we eliminate the sparsity constraint on the rows of the coefficient matrix {\small$\mb{A}$} and the orthogonality constraint on the rows of the dictionary {\small$\mb{R}$} required by \cite{Mardani2012}.} 

	 %
	 %
	 %
	 %
	 
	The model shown in eq.\eqref{Prob} is propitious in a number of applications.  For example, it  can be used for target identification in hyperspectral imaging, and in topic modeling applications to identify documents with certain properties. Further, in source separation tasks, a variant of this model was used in singing voice separation in \cite{Huang2012, Sprechmann2012}.  Further, we can also envision source separation tasks where {\small$\mb{X}$} is not low-rank, but can in turn be modeled as being sparse in a known \cite{Starck2005} or unknown \cite{Rambhatla2013} dictionary. 
	
	The rest of the paper is organized as follows. We formulate the problem, introduce the notation and describe various considerations on the structure of the component matrices in section~\ref{Formulation}. In section~\ref{MainResult}, we present our main result and a proof sketch, followed by numerical simulations in section~\ref{Simulations}. Finally, we conclude in section~\ref{Conclusion} with some insights on future work. 
	\vspace{-10pt}
	\section{Problem Formulation}
	\vspace{-8pt}
	\label{Formulation}	
 	Our aim is to recover the low-rank component {\small$\mb{X}$}, and the sparse coefficient matrix {\small$\mb{A}$}, given the dictionary {\small$\mb{R}$}, and samples {\small$\mb{Y}$} generated according to the model described in eq.\eqref{Prob}. Utilizing the \edit{assumed} structure of the components {\small$\mb{X}$} and {\small$\mb{A}$}, we consider the following convex problem for {\small$\lambda \geq 0 $},
	\startcompact{small}
	\begin{align}
	\label{optProb}
	\underset{\mb{X}, \mb{A}}{\text{minimize~}} \|\mb{X}\|_* + \lambda \|\mb{A}\|_1 ~~\text{s.t.}~~ \mb{Y} = \mb{X} + \mb{RA}.
	\end{align}
	\stopcompact{small}%
	where, {\small$\|.\|_* $} denotes the nuclear norm, and {\small$\|.\|_1$} refers to the $l_1$- norm, which serve as convex relaxations of rank and sparsity (i.e. $l_0$-norm), respectively. Depending upon the number of dictionary elements, {\small$d$} in {\small$\mb{R}$}, we analyze the problem described above for two cases -- a) when  {\small $d \leq n$}, the \textit{thin} case, and b) when {\small $d > n$}, the \textit{fat} case. \edit{For the \textit{thin} case, we assume that the rows of the dictionary {\small$\mb{R}$} to comprise a \textit{frame}, i.e. for any vector $\mb{v}\in \mathbb{R}^{d}$, we have}
	\startcompact{small}
	\begin{align}
	\label{frame}
	\mb{F}_{L}\|\mb{v}\|^2_2 \leq \|\mb{Rv}\|^2_2 \leq \mb{F}_U\|\mb{v}\|^2_2, 
	\end{align}
	\stopcompact{small}%
	where {\small$\mb{F}_{L}$} and {\small$\mb{F}_U$} are the lower and upper \textit{frame bounds}, respectively, with {\small$0<\mb{F}_{L}\leq\mb{F}_U$}. 
	Next, for {\small $d > n$}, the \textit{fat} case, we assume that {\small$\mb{R}$} obeys the restricted isometry property (RIP) of order {\small $k$}, i.e. for any {\small $k$}-sparse vector $\mb{v}\in \mathbb{R}^d$, we have
	\startcompact{small}
	\begin{align}
		\label{rip}
		(1 - \delta)\|\mb{v}\|^2_2 \leq \|\mb{Rv}\|^2_2 \leq (1 + \delta)\|\mb{v}\|^2_2, 
	\end{align}
	\stopcompact{small}%
	where, {\small $\delta \in [0, 1]$} is the restricted \edit{isometry} constant (RIC). 
		
	The aim of this paper is to answer the following question -- Given {\small$\mb{R}$}, under what conditions can we recover {\small$\mb{X}$} and {\small$\mb{A}$} from the mixture {\small$\mb{Y}$}? 
	We observe that there are a few ways we can run into trouble right away, namely-- a) the dictionary sparse part {\small$\mb{RA}$} is low-rank, and  b) the low rank part, {\small$\mb{X}$}, is sparse in the dictionary, {\small$\mb{R}$}. Indeed, these relationships take the center stage in our analysis. We begin by defining a few relevant subspaces, similar to those used in \cite{Mardani2012}, which will help us to formalize the said relationships. 
	
	Let the pair {\small$\{\mb{X_0, A_0} \}$} be the solution to the problem shown in eq.\eqref{optProb}. We define {\small$\mb{\Phi}$} as the linear space of matrices spanning the row and column spaces of the low-rank component {\small$\mb{X}_0$}. Specifically, let {\small$\mb{U \Sigma V}'$} denote the singular value decomposition of {\small$\mb{X}_0$}, then  the space  {\small$\mb{\Phi}$} is defined as
	\startcompact{small}
	\begin{align*}
	\mb{\Phi} := \{ \mb{UW}_1 + \mb{W}_2\mb{V}', \mb{W}_1, \mb{W}_2 \in \mathbb{R}^{n \times r}\}.
	\end{align*}
	\stopcompact{small}%
	Next, let {\small$\mb{\Omega}$} be the space \edit{spanned by {\small ${d \times m}$} matrices that} have the same support (location of non-zero elements) as {\small$\mb{A}_0$}, and let {\small$\mb{\Omega_R}$} be defined as
 	\startcompact{small}
 	\begin{align*}
 	\mb{\Omega_R} := \{ \mb{Z} = \mb{RH}, \mb{H} \in \mb{\Omega} \}.
 	\end{align*}
 	\stopcompact{small}%
 	 In addition, we denote the corresponding complements of the spaces described above by appending `$\perp$'.  Next, let the orthogonal projection operator(s) onto the space of matrice(s) defined above be {\small$\mc{P}_\Phi(.)$}, {\small$\mc{P}_\Omega(.)$} and {\small$\mc{P}_{\Omega_R}(.)$}, respectively. Further, we will use {\small$\mb{P}_U$} and {\small$\mb{P}_V$} to denote the projection matrices corresponding to the column and row spaces of {\small$\mb{X}_0$}, respectively, i.e., implying the following for any matrix {\small$\mb{X}$},%
    \startcompact{small}
	\begin{align*}
	\mc{P}_\Phi (\mb{X}) &= \mb{P}_U\mb{X} + \mb{X} \mb{P}_V -   \mb{P}_U\mb{X}\mb{P}_V\\
	\mc{P}_{\Phi^\perp}(\mb{X}) &=  (\mb{I} - \mb{P}_U)\mb{X}  (\mb{I} - \mb{P}_V).
	\end{align*}
	\stopcompact{small}%
	As alluded to previously, there are indeed situations under which we cannot hope to recover the matrices {\small$\mb{X}$} and {\small$\mb{A}$}. To identify these scenarios, we will employ various notions of incoherence. \edit{We define} the incoherence between the low-rank part, {\small$\mb{X}_0$}, and the dictionary sparse part, {\small $\mb{RA_0}$} as, 
	\startcompact{small}
	\begin{align*}
	\mu := \underset{\mb{Z} \in \mb{\Omega}_R \backslash \{\mb{0}_{d \times m}\}}{\text{max}} \tfrac{\|\mathcal{P}_{\Phi}(\mb{Z})\|_F}{\|\mb{Z}\|_F},
	\end{align*}
	\stopcompact{small}%
	where {\small $\mu \in [0, 1]$} is small when these components are incoherent (good for recovery).  The next two measures of incoherence can be interpreted as a way to avoid the cases where for {\small$\mb{X}_0 = \mb{U \Sigma V'}$}, (a) {\small$\mb{U}$} resembles the dictionary  {\small$\mb{R}$}, and (b) {\small$\mb{V}$} resembles the sparse coefficient matrix {\small$\mb{A}_0$}. In this case, the low-rank part essentially mimics the dictionary sparse component. To this end, similar to \cite{Mardani2012}, we define respectively the following to measure these properties,
	\startcompact{small}
    \begin{align*}
	\mb{\gamma}_{U\!R}  :=  \underset{i}{\text{max}} \tfrac{\|\mb{P}_U \mb{R}\mb{e}_{i}\|^2}{\|\mb{Re}_{i}\|^2} ~{\normalsize\text{and}}~
	\mb{\gamma}_V := \underset{i}{\text{max}} \|\mb{P}_V\mb{e}_{i}\|^2,
	\end{align*}
	\stopcompact{small}%
	 where {\small$\mb{\gamma}_V  {\small\in [r/m, 1]}$}. Also, we define {\small$\xi := \|\mb{R}'\mb{UV}'\|_\infty$}.
	%
\vspace{-10pt}
\section{Main Result}
\label{MainResult}
\vspace{-8pt}
In this section, we present the conditions under which solving the problem stated in eq.\eqref{optProb} will successfully recover the true matrices {\small$\mb{X}$} and {\small$\mb{A}$}. As discussed in the previous section, the structure of the dictionary {\small$\mb{R}$} plays a crucial role in the analysis of the two paradigms, i.e. the \textit{thin} case and the \textit{fat} case. Consequently, we provide results corresponding to these cases separately. We begin by introducing a few definitions and assumptions applicable to both cases. To simplify the analysis we assume that {\small$d < m$}. \edit{Specifically, we will assume that {\small$d \leq \tfrac{m}{\alpha r}$}, where {\small$\alpha >1$} is a constant. In addition, our analysis is applicable to the case when {\small$s < m$}.}
\vspace{-5pt}
\begin{definition}
\label{lamMin}
{\small We define} \vspace{-4pt}
\startcompact{footnotesize}
\begin{align}
c := {\scriptsize
	\begin{cases}
     c_t , \text{for  $ d \leq n$}\\
     c_f , \text{for  $ d > n$} 
     \end{cases}}, \notag \end{align}
\stopcompact{footnotesize}%
{\small where, {\scriptsize $c_t$ } and {\scriptsize $c_f$ } are defined as,} \\
{\scriptsize $c_t := \tfrac{\mb{F}_U}{2} [(1 + 2\gamma_{U\!R} )(\text{min}(s, d)  + s\gamma_V ) +2s\gamma_V] - \tfrac{\mb{F}_L}{2}[\text{min}(s, d)  + s\gamma_V ]$~and} \\ {\scriptsize $c_f := \tfrac{1 + \delta}{2}[(1 + 2\gamma_{U\!R} )(\text{min}(s, k)  + s\gamma_V ) +2s\gamma_V] - \tfrac{1 - \delta}{2}(\text{min}(s, k)  + s\gamma_V ) $}. 
\noindent Further, we define {\small $C$} and {\small $\lambda_{\text{min}}$} as,
\startcompact{small}
\begin{align}
{\small C} := \hspace{-2pt}{\scriptsize
	\begin{cases}
     \tfrac{c_t}{\mb{F}_L(1 - \mu)^2 - c_t}, & \hspace{-5pt}\text{for  $ d \leq n$ and ${\scriptsize \mb{F}_L \leq \tfrac{1}{(1 - \mu)^2}}$}\\
     \tfrac{c_f}{(1 - \delta)(1 - \mu)^2 - c_f} \hspace{-2pt}, & \hspace{-5pt} \text{for  $ d > n$} 
     \end{cases}} \notag  \hspace{-2pt} \text{\small~and~}\lambda_{\text{min}}  \hspace{-2pt}:= \tfrac{1 + C}{1-C} ~\xi.
 \end{align}
\stopcompact{small}%
\end{definition}
\vspace{0.05em}
\begin{definition}%
\label{lamMax}%
\vspace{-5pt}
	\startcompact{scriptsize}%
	\begin{align*}%
	\lambda_{\text{max}} := 
	{\scriptsize\begin{cases}
	\tfrac{1}{\sqrt{s}} \big(\sqrt{\mb{F}_{L}} {(}~ 1- \mu ~{)} -\sqrt{r \mb{F}_U} \mu\big{)}, &\mbox{if } d \leq n \\
	\tfrac{1}{\sqrt{s}} \big(\sqrt{(1 - \delta)} {(}~ 1- \mu ~{)} -\sqrt{r (1 + \delta)} \mu\big{)}, & \mbox{if } d > n
	\end{cases} }
	\end{align*}
	\stopcompact{scriptsize}%
	\end{definition}
\begin{assumption}
\label{A1}
$\lambda_{\text{max}} \geq \lambda_{\text{min}}$
\vspace{-5pt}
\end{assumption}
\vspace{-3pt}
\begin{assumption}
\label{A2}
\startcompact{small}%
Let $s_{\text{max}} := \tfrac{(1 - \mu)^2}{2}\tfrac{m}{r} $, then
\begin{align*}%
\gamma_{U\!R} ~~\leq ~~
	\begin{cases}
     \tfrac{(1 - \mu)^2 - 2s\gamma_V}{2s( 1 + \gamma_V)}, \text{ for }     s \leq \text{\small min} ~(d, s_{\text{max}})\\
     \tfrac{(1 - \mu)^2 - 2s\gamma_V}{2(d + s\gamma_{V})}, \text{ for }   d<s \leq s_{\text{max}} 
     \end{cases}. \notag 
\end{align*}
\stopcompact{small}%
\end{assumption}
\vspace{-5pt}
\begin{assumption}
\label{A4}
\startcompact{small}%
For {\small$s_{\text{max}}$} as above,
\begin{align*}%
\gamma_{U\!R} ~~\leq ~~
	\begin{cases}
     \frac{(1 - \mu)^2 - 2s\gamma_V}{2s( 1 + \gamma_V)}, \text{ for }    s \leq \text{\small min} ~(k, s_{\text{max}})\\
     \frac{(1 - \mu)^2 - 2s\gamma_V}{2(k + s\gamma_{V})}, \text{ for }   k<s \leq s_{\text{max}} 
     \end{cases}. \notag 
\end{align*}
\stopcompact{small}%
\end{assumption}
\vspace{-10pt}

\vspace{8pt}
\startcompact{small}
\begin{theorem}
\label{theorem}
Consider a superposition of a low-rank matrix {\small $\mb{X}_0 \in \mathbb{R}^{n \times m}$} of rank {\small $r$}, and a dictionary sparse component {\small $\mb{RA_0}$}, wherein %
the sparse coefficient matrix {\small $\mb{A}_0$} has at most {\small $s$} non-zeros, i.e., {\small $\|\mb{A}_0\|_0 = s$}, and {\small $\mb{Y} = \mb{X}_0 + \mb{RA_0}$}, with parameters {\small$\gamma_{U\!R}$},  {\small $\xi$}, {\small$\mb{\gamma}_V \in [r/m, 1]$} and  {\small$\mu \in [0, 1]$}. Then, solving the formulation shown in eq.\eqref{optProb} will recover matrices {\small $\mb{X}_0$} and {\small $\mb{A}_0$} if the following conditions hold for any {\small $\lambda \in [\lambda_{\text{min}}, \lambda_{\text{max}} ]$}, as defined in D.\ref{lamMin} and  D.\ref{lamMax}, respectively.

$\bullet$ For $d\leq n$, the dictionary {\small $\mb{R} \in \mathbb{R}^{n \times d}$} obeys the frame condition with frame bounds {\small $[\mb{F}_{L}, \mb{F}_U ]$} and assumptions A.\ref{A1}, and A.\ref{A2} hold.

$\bullet$ For $d> n > C_1k \text{log}(d)$, the dictionary {\small $\mb{R} \in \mathbb{R}^{n \times d}$} obeys the RIP of order {\small $k$} with RIC {\small $\delta$}, a constant {\small $C_1$} and assumptions A.\ref{A1} and A.\ref{A4} hold.
\vspace{-6pt}
\end{theorem}
\stopcompact{small}
Thm.~\ref{theorem} establishes the sufficient conditions for the existence of {\small $\lambda$}s to guarantee recovery of  {\small$\{\mb{X_0, A_0} \}$} for both the \textit{thin} and the \textit{fat} case. For both cases, we see that the conditions are closely related to the various incoherence measures {\small$\gamma_{U\!R}$}, {\small$\mb{\gamma}_V$} and {\small$\mu$} between the low-rank part, {\small$ \mb{X}$}, the dictionary, {\small$ \mb{R}$}, and the sparse component {\small$ \mb{A}$}. Further, we observe that the theorem imposes an an upper-bound on the global sparsity, i.e., $s \leq s_{\text{max}}$. This is similar to what was reported in \cite{Xu2010}, and seems a result of the deterministic analysis presented here. 
%
%
%
%
%
Further, the condition shown in assumption A.\ref{A1}, i.e., {\small$\lambda_{\text{min}} \leq \lambda_{\text{max}}$}, translates to a relationship between rank {\small $r$}, and the sparsity {\small $s$}, namely,

\startcompact{small}
\begin{align}
\label{rankSpar1}
r \leq& \bigg(\sqrt{\tfrac{\mb{F}_{L}}{\mb{F}_{U}}} \tfrac{ 1-\mu}{\mu} - \tfrac{\xi}{\sqrt{\mb{F}_{U}}\mu} \tfrac{1 + C}{1 - C} \sqrt{s}\bigg)^2,
\end{align}
\stopcompact{small}%
{\small$\forall s \geq 0 ~{\normalsize\text{such that}}, ~\sqrt{s} \leq \tfrac{\mb{F}_{L} ( 1- \mu)}{\xi} \tfrac{1 - C}{1 + C}$}, for the \textit{thin} case, and  
\startcompact{small}
\begin{align}
\label{rankSpar2}
r \leq& \bigg(\sqrt{\tfrac{1 - \delta}{1 + \delta}} \tfrac{ 1-\mu}{\mu} - \tfrac{\xi}{\sqrt{(1 + \delta)}\mu} \tfrac{1 + C}{1 - C} \sqrt{s}\bigg)^2,
\end{align}
\stopcompact{small}%
{\small$\forall s \geq 0 ~{\normalsize\text{such that}}, ~\sqrt{s} \leq \tfrac{(1 - \delta) ( 1- \mu)}{\xi} \tfrac{1 - C}{1 + C}$}, for the \textit{fat} case.
%
These relationships are indeed what we observe in empirical evaluations; this will be revisited in the next section. We now present a brief proof sketch of the results presented in this section. 
\vspace{-12pt}
\subsection{Proof Sketch}
\vspace{-5pt}
We use dual certificate construction procedure to prove the main result in Thm.~\ref{theorem} \cite{RambhatlaPrep}. To this end, we start by constructing a dual certificate for the convex problem shown in eq.\eqref{optProb}. In our analysis, we use {\small$\|\mb{M}\|:= 
	\sigma_{\text{max}}(\mb{M})$} for the spectral norm, here {\small$\sigma_{\text{max}}(\mb{M})$} denotes the maximum singular value of the matrix {\small$\mb{M}$}, {\small$\|\mb{M}\|_\infty := \underset{\{i,~j\}}{\text{max}} |M_{ij}|$}, and {\small$\|\mb{M}\|_{\infty, \infty} := 
	 \underset{i}{\text{max}} \|\mb{e}'_i\mb{M}\|_1$}. 	
	 The following lemma shows the conditions the dual certificate needs to satisfy.
	 \vspace{-5pt}
	{\small 
	\begin{lemma}[from Lemma 2 in \cite{Mardani2012} and Thm. 3 in \cite{Xu2010}]\label{DualCert}
	\textit{If there exists a dual certificate {\small$\mb{\Gamma} \in \mathbb{R}^{n \times m }$} satisfying} 
	\startcompact{footnotesize}
	\begin{align*}
	\centering
	 \begin{tabular}{ll}
	\text{C}1 : $\mc{P}_\Phi(\mb{\Gamma})= \mb{UV'}$ &
	\text{C}2 : $\mc{P}_\Omega(\mb{R'\Gamma})= \lambda \text{sign(}\mb{A}_0\text{)}$\\
	\text{C}3 : $\|\mc{P}_{\Phi^\perp} (\mb{\Gamma})\| ~<~ 1$ &
	\text{C}4 : $\|\mc{P}_{\Omega^\perp} (\mb{R'\Gamma})\|_\infty ~<~ \lambda$ \\
	\end{tabular}
	\end{align*}
	\stopcompact{footnotesize}%
	\textit{then the pair {\small$\{ \mb{X}_0,~\mb{A}_0\}$} is the unique solution of eq~(\ref{optProb}).}
	\end{lemma}} \vspace{-5pt}

We will now proceed with the construction of the dual certificate which satisfies the conditions outlined by conditions \text{C}1-4 by Lemma~\ref{DualCert}. Using the analysis similar to \cite{Mardani2012} (section V. B.), we construct the dual certificate as 
\startcompact{small}
\begin{align*}
\mb{\Gamma} =\mb{UV'} + (\mb{I- P_U})\mb{X} \mb{(I - P_V)},
\end{align*}
\stopcompact{small}%
for arbitrary {\small$\mb{X} \in \mathbb{R}^{n\times m}$}. The condition C2 then translates to 
\startcompact{small}
\begin{align*}	
\mc{P}_\Omega(\mb{R'UV'}) + \mc{P}_\Omega(\mb{R'(I- P_U})\mb{X} \mb{(I - P_V)})&= \lambda ~\text{sign}(\mb{A}_0) 
\end{align*}
\stopcompact{small}%
Let {\small$\mb{Z} := \mb{R'}(\mb{I- P_U})\mb{X} \mb{(I - P_V)}$} and {\small$\mb{B_\Omega} :=\lambda \text{sign}(\mb{A}_0) - \mathcal{P}_\Omega (\mb{R'UV'})$}, then we can write the equation above as,
\startcompact{small}
\begin{align*}
			\mc{P}_\Omega(\mb{Z}) &= \mb{B_\Omega}.
\end{align*}
\stopcompact{small}%
Note that {\small$\text{vec}(\mb{Z}) = [\mb{(I - P_V)}\otimes \mb{R'}(\mb{I- P_U})] \text{vec}(\mb{X})$}. Now, let {\small$\mb{\tilde{A}} := \mb{(I - P_V)}\otimes \mb{R'}(\mb{I- P_U})$}, and let {\small$\mb{\tilde{A}_\Omega} \in \mathbb{R}^{s \times nm}$} denote the rows of {\small$\mb{\tilde{A}}$} that correspond to support of {\small$\mb{A}_0$}, and {\small$\mb{\tilde{A}_{\Omega^\perp}}$} correspond to the remaining rows of {\small$\mb{\tilde{A}}$}. Further, let  {\small$\mb{b}_\Omega$} be a length {\small $s$} vector containing elements of  {\small$\mb{B}_\Omega$} corresponding to support of  {\small$\mb{A}_0$}. Using these definitions and results, we conclude
\vspace{-10pt}
\startcompact{small}
\begin{align*}
\mb{\tilde{A}}_\Omega \text{vec}(\mb{X}) = \mb{b}_\Omega
\end{align*}
\stopcompact{small}%
This implies that {\small$\text{vec}(\mb{X}) = \mb{\tilde{A}}_\Omega'(\mb{\tilde{A}}_\Omega\mb{\tilde{A}}_\Omega')^{-1}\mb{b}_\Omega$}.
Now, we look at the condition C3, i.e. {\small$\|\mathcal{P}_{\Phi^\perp}(\mb{\Gamma})\|$}, this is where our analysis departs from \cite{Mardani2012}; we write
\vspace{-0.5cm}
	\startcompact{footnotesize}
	\begin{align*}
	\\
	\|\mathcal{P}_{\Phi^\perp}(\mb{\Gamma})\| &= \|(\mb{I- P_U})\mb{X} \mb{(I - P_V)}\|
	\leq \|(\mb{I- P_U})\|\|\mb{X}\|\| \mb{(I - P_V)}\|\\ &\leq \|\mb{X}\| \leq \|\mb{X}\|_F
	 \leq \|\mb{\tilde{A}}_\Omega' (\mb{\tilde{A}}_\Omega\mb{\tilde{A}}_\Omega')^{-1}\| \|\mb{b}_\Omega\|_2,
	\end{align*}
	\stopcompact{footnotesize}%
	where we have used the fact that {\small$\|(\mb{I- P_U})\| \leq 1$} and {\small $\| \mb{(I - P_V)}\| \leq 1$}. Now, as {\small$\mb{\tilde{A}}_\Omega' (\mb{\tilde{A}}_\Omega\mb{\tilde{A}}_\Omega')^{-1}$} is the pseudo-inverse of {\small $\mb{\tilde{A}}_\Omega$, i.e., $\mb{\tilde{A}}_\Omega\mb{\tilde{A}}_\Omega' (\mb{\tilde{A}}_\Omega\mb{\tilde{A}}_\Omega')^{-1} = \mb{I}$}, we have that {\small$\|\mb{\tilde{A}}_\Omega' (\mb{\tilde{A}}_\Omega\mb{\tilde{A}}_\Omega')^{-1}\| = 1/{\sigma_{\text{min}}{(\mb{\tilde{A}}_\Omega)}}$}, where { \small$\sigma_{\text{min}}{(\mb{\tilde{A}}_\Omega)}$} is the smallest singular value of {\small $\mb{\tilde{A}}_\Omega$}. Therefore, we have
\startcompact{small}
	\begin{align*}
	\|\mathcal{P}_{\Phi^\perp}(\mb{\Gamma})\| & \leq \tfrac{\|\mb{b}_{\Omega}\|_2}{\sigma_{\text{min}}{(\mb{\tilde{A}}_\Omega)}} .
	\end{align*}
	\stopcompact{small}%
	To obtain an upper bound on {\small$\|\mathcal{P}_{\Phi^\perp}(\mb{\Gamma})\| $}, we will now present the following lemmata.%
	 \vspace{-8pt}
	{\small
	\begin{lemma}\label{lower_sigmaMin}
	\textit{The lower bound on {\small$\sigma_{\text{min}}{(\mb{\tilde{A}}_\Omega)}$} is given by} 
	\startcompact{footnotesize}
	\begin{align}
	\sigma_{\text{min}}{(\mb{\tilde{A}}_\Omega)} \geq {\scriptsize
		\begin{cases}
	     \sqrt{\mb{F}_{L}} \big{(}~ 1- \mu ~\big{)} , \text{for  $ d \leq n$}\\
	     \sqrt{(1 - \delta)} \big{(}~ 1- \mu ~\big{)} , \text{for  $ d > n$} 
	     \end{cases}}. \notag \end{align}
	\stopcompact{footnotesize}%
	\end{lemma}}
	 \vspace{-12pt}
	{\small
	\begin{lemma}\label{upper_bOmega}
	 	\textit{	Upper bound on {\small$\|\mb{b}_\Omega\|_2$} is given by}  
	 \startcompact{footnotesize}
	 	\begin{align}
	 \|\mb{b}_\Omega\|_2 \leq {\scriptsize
	 		\begin{cases}
	 	     \lambda \sqrt{s} + \sqrt{r \mb{F}_U} \mu , \text{for  $ d \leq n$}\\
	 	     \lambda \sqrt{s} + \sqrt{r (1 + \delta)} \mu , \text{for  $ d > n$} 
	 	     \end{cases}}. \notag \end{align}
	 	\stopcompact{footnotesize}%
	 	\end{lemma}}
	 	\vspace{-5pt}%
	 Assembling the results of the lemmata to obtain the upper bound on {\small$\|\mathcal{P}_{\Phi^\perp}(\mb{\Gamma})\|$} and consequently C3, we arrive at the expression for {\small$\lambda_{\text{max}}$} defined in D.\ref{lamMax}.
Now, we move onto finding conditions under which C4 is satisfied by our dual certificate. For this we will bound {\small $\|\mc{P}_{\Omega^\perp} (\mb{R'\Gamma})\|_\infty $}. From eq.(16) in \cite{Mardani2012} we have, 
	\startcompact{small}
\begin{align}
\label{C3exp}
\|\mc{P}_{\Omega^\perp} (\mb{R'\Gamma})\|_\infty  \leq \|\mb{Q}\|_{\infty, \infty} \|\mb{b}_\Omega\|_\infty + \|\mc{P}_{\Omega^\perp}(\mb{R'UV'})\|_\infty,
\end{align}
\stopcompact{small}%
where 	{\small$\mb{Q} : =\mb{\tilde{A}}_{\Omega^\perp} \mb{\tilde{A}}_\Omega' (\mb{\tilde{A}}_\Omega\mb{\tilde{A}}_\Omega')^{-1}$}. Our aim now will be to bound {\small$\|\mb{Q}\|_{\infty, \infty}$} and {\small$\|\mb{b}_\Omega\|_\infty$} for our case. For this, we present the following lemmata.
\vspace{-8pt}
{\small
\begin{lemma}[from eq.(17) in \cite{Mardani2012} ]\label{lbOmegaInf}
\textit{The upper bound on {\small$\|\mb{b}_\Omega\|_\infty$} is given by {\small$ \lambda  + \|\mc{P}_\Omega (\mb{R}'\mb{UV}')\|_\infty$.}}
\end{lemma}}\vspace{-8pt}
\vspace{-8pt}
{\small
\begin{lemma}\label{Q_inf}
\textit{The upper bound on {\small$\|\mb{Q}\|_{\infty, \infty}$} is given by  {\small$C$}, where {\small$C$} is as defined in D.\ref{lamMin}.}
\end{lemma}}\vspace{-8pt}
%
\noindent Substituting these in eq.\eqref{C3exp} and C4, we have 
\startcompact{small}
\begin{align*}
\|\mc{P}_{\Omega^\perp} (\mb{R'\Gamma})\|_\infty \leq C (\lambda  + \|\mc{P}_\Omega (\mb{R}'\mb{UV}')\|_\infty) + \|\mc{P}_{\Omega^\perp}(\mb{R'UV'})\|_\infty.
\end{align*}
\stopcompact{small}%
The expression above is further upper bounded by {\small$\lambda$} due to C4. Here, {\small$C$} and {\small$c$} are as defined in D.\ref{lamMin}. Hence, we arrive at the following  lower bound for {\small$\lambda$},
\startcompact{footnotesize}
\begin{align*}
      \lambda_{\text{min}} := \dfrac{1 + C}{1-C} ~\xi.
\end{align*}
\stopcompact{footnotesize}%
Gleaning from the expressions for {\small$\lambda_{\text{max}}$} and {\small$\lambda_{\text{min}}$}, we observe that the following conditions need to be satisfied for the existence of {\small$\lambda$}s that can recover the desired matrices-- a) {\small$\lambda_{\text{max}} \geq \lambda_{\text{min}} >0$}, and b) {\small$0<C<1$}. These conditions are satisfied by the assumptions A.\ref{A1}-A.\ref{A4}. This completes the proof. 
\vspace{-10pt}
\section{Simulations}
\label{Simulations}
\begin{figure}[h]
    	\centering
    			{\small
    				  \begin{tabular}{cc}
    				  \vspace{-0.1cm}
    				      ~~~~$d = 5$ & ~~~~$d = 150$\\
    				    \hspace{-0.2cm} {\small\rotatebox{90}{ ~~~~~~~~~Recovery of $\mb{X}$}}
    				     \includegraphics[width=0.225\textwidth]{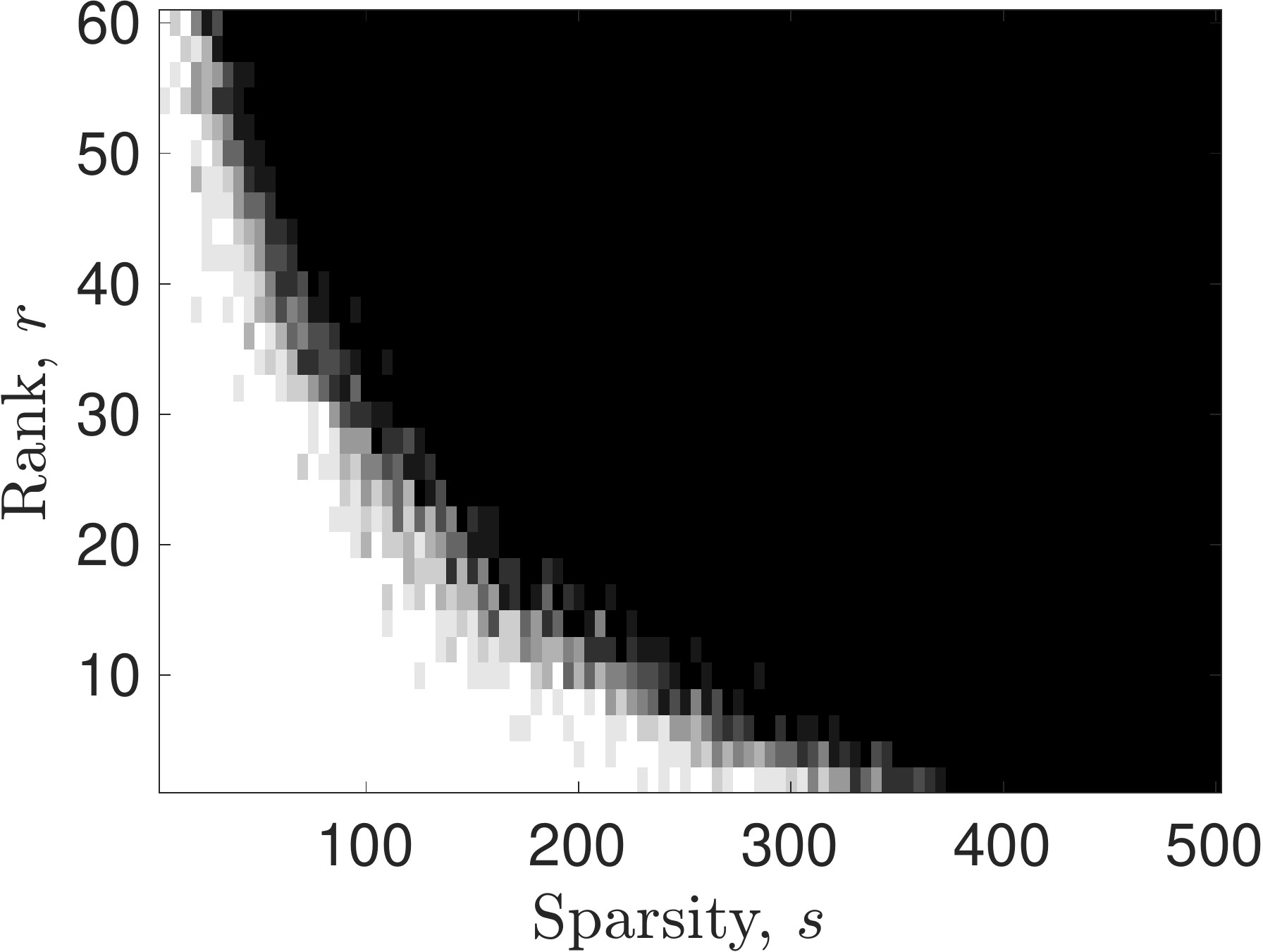} &\hspace{-0.4cm} \includegraphics[width=0.22\textwidth]{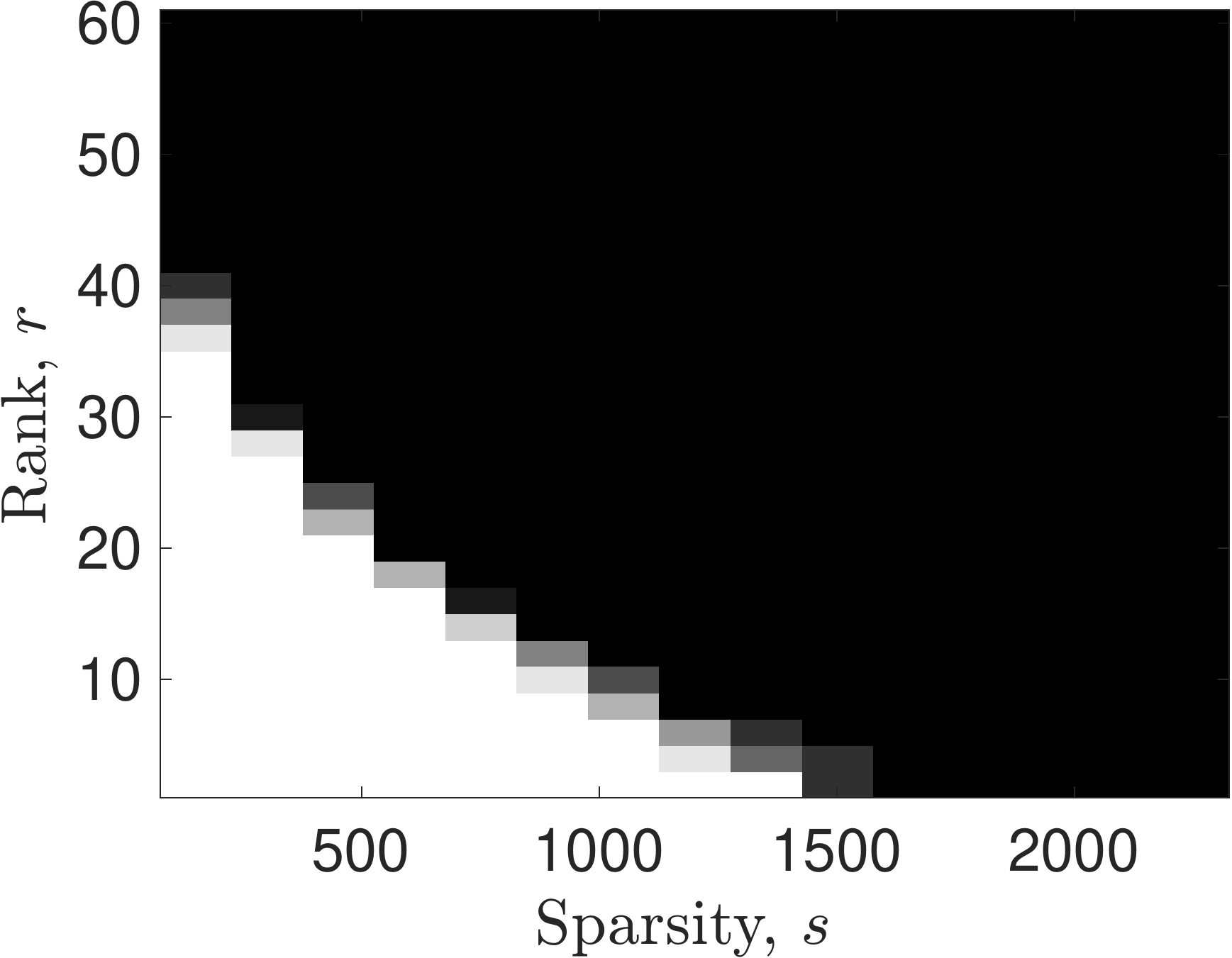} \vspace{-4pt}\\
    				     ~~~~~~~~(a) & ~~(b) \vspace{-1pt}\\
    				     \hspace{-0.2cm}{\small\rotatebox{90}{ ~~~~~~~~~Recovery of $\mb{A}$}}
    				     \includegraphics[width=0.225\textwidth]{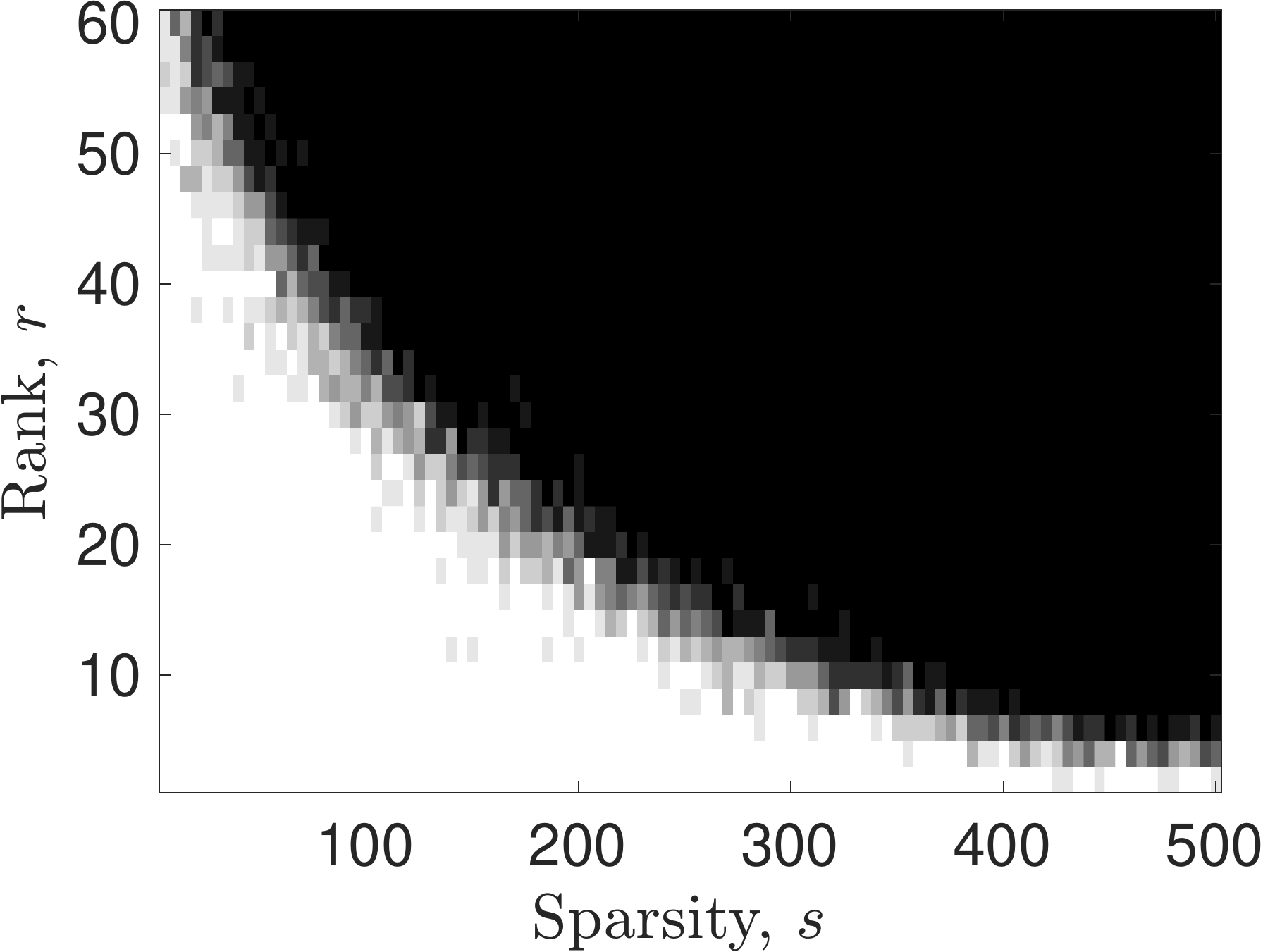} & \hspace{-0.4cm} \includegraphics[width=0.22\textwidth]{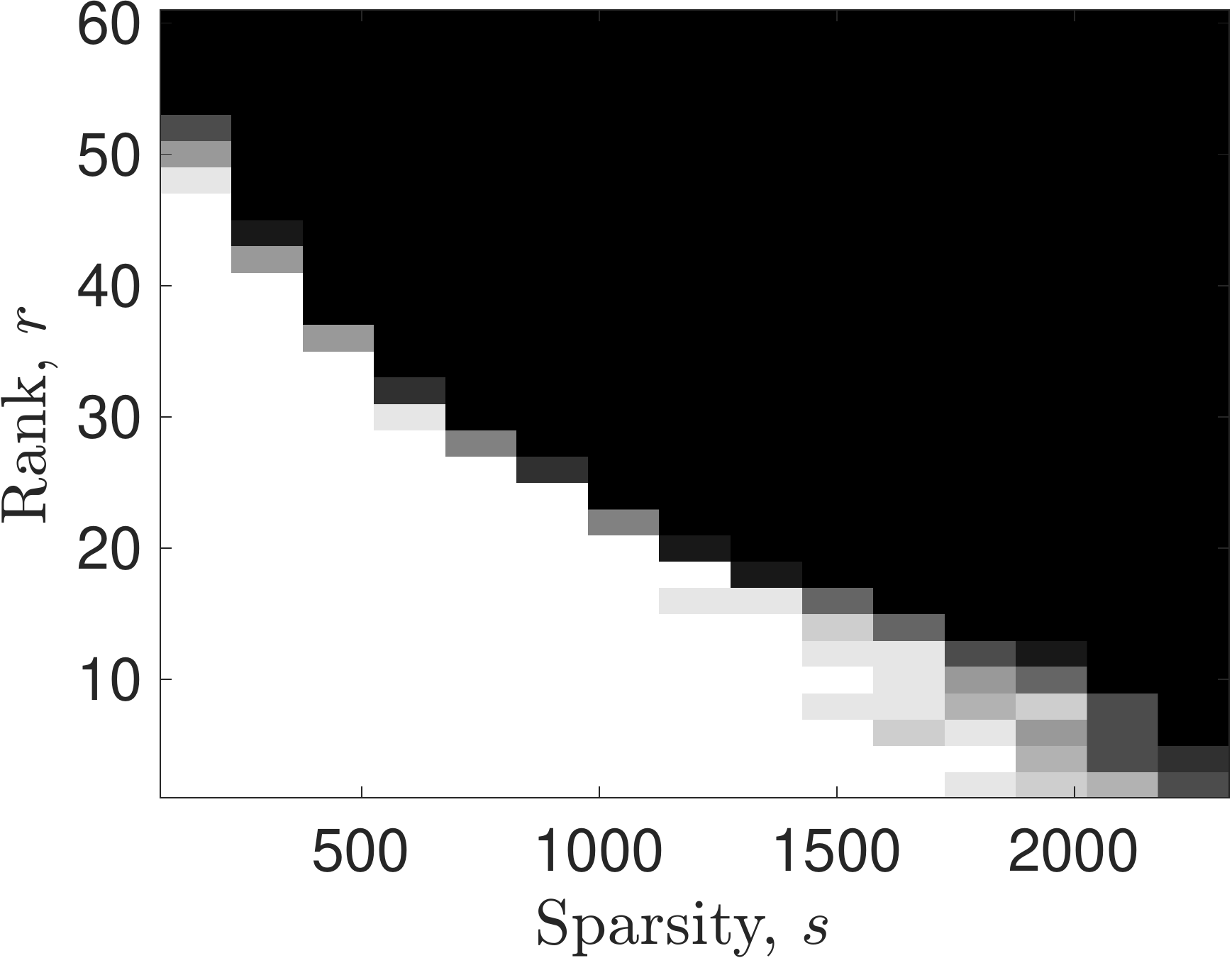} \vspace{-4pt}\\
    				    ~~~~~~~~(c) & ~~(d)   
    				     \end{tabular}}
    				     \vspace{-0.4cm}
    				     \caption{Recovery for varying rank of {\footnotesize$\mb{X}$}, sparsity of {\footnotesize $\mb{A}$} and number of dictionary elements in {\footnotesize $\mb{R}$}. Each plot shows average recovery across 10 trials for varying ranks (y-axis) and sparsity (x-axis), white region representing correct recovery, for $n = m = 100$. We decide success if {\footnotesize $\| \mb{X} - \hat{\mb{X}}\|_F/\| \mb{X} \|_F \leq 0.02$} and {\footnotesize $\| \mb{A} - \hat{\mb{A}}\|_F/\| \mb{A} \|_F \leq 0.02$}, where {\footnotesize $\hat{\mb{X}}$} and {\footnotesize $\hat{\mb{A}}$} are the recovered {\footnotesize $\mb{X}$} and {\footnotesize $\mb{A}$}, respectively. Panels (a)-(b) show the recovery of the low-rank part {\footnotesize $\mb{X}$} and (c)-(d) show the recovery of the sparse part with varying dictionary sizes $d = 5 ~\text{and}~ 150$, respectively.}
    				     \vspace{-0.7cm}
    				     \label{fig:phaseTr_full}
    \end{figure}

\startcompact{footnotesize}
\begin{figure}[h]
\centering
	{\small
		  \begin{tabular}{cc}
		  \vspace{-0.1cm}
		      ~~~~$d = 5$ & ~~~~$d = 150$\\
		    \hspace{-0.2cm} {\small\rotatebox{90}{ ~~~~~~~~~Recovery of $\mb{X}$}}
		     \includegraphics[width=0.225\textwidth]{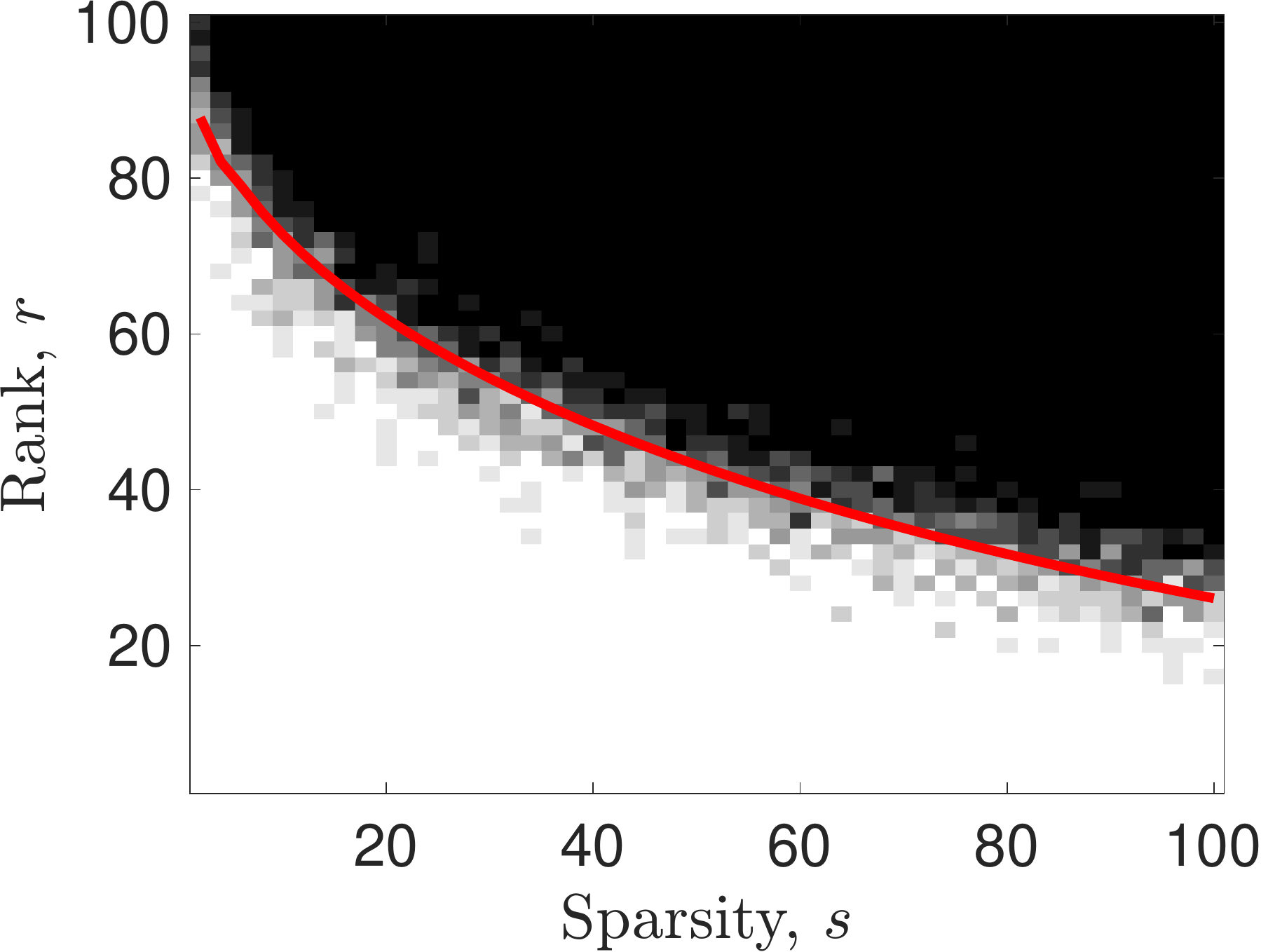} & \hspace{-0.4cm} \includegraphics[width=0.225\textwidth]{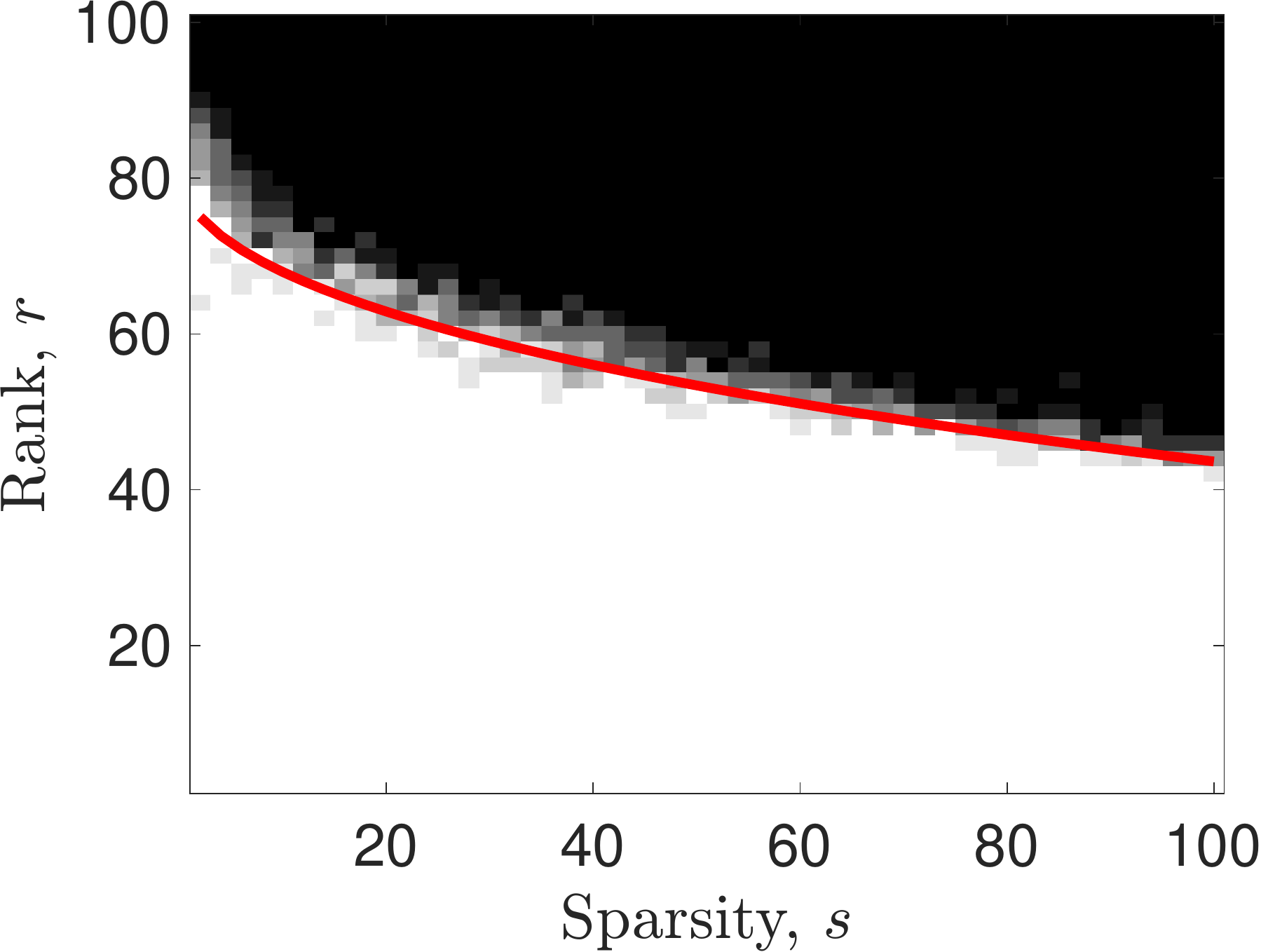} \vspace{-4pt}\\
		     ~~~~~~~~(a) & ~~(b)  \vspace{-1pt}\\
		      \hspace{-0.2cm}{\small\rotatebox{90}{ ~~~~~~~~~Recovery of $\mb{A}$}}
		     \includegraphics[width=0.225\textwidth]{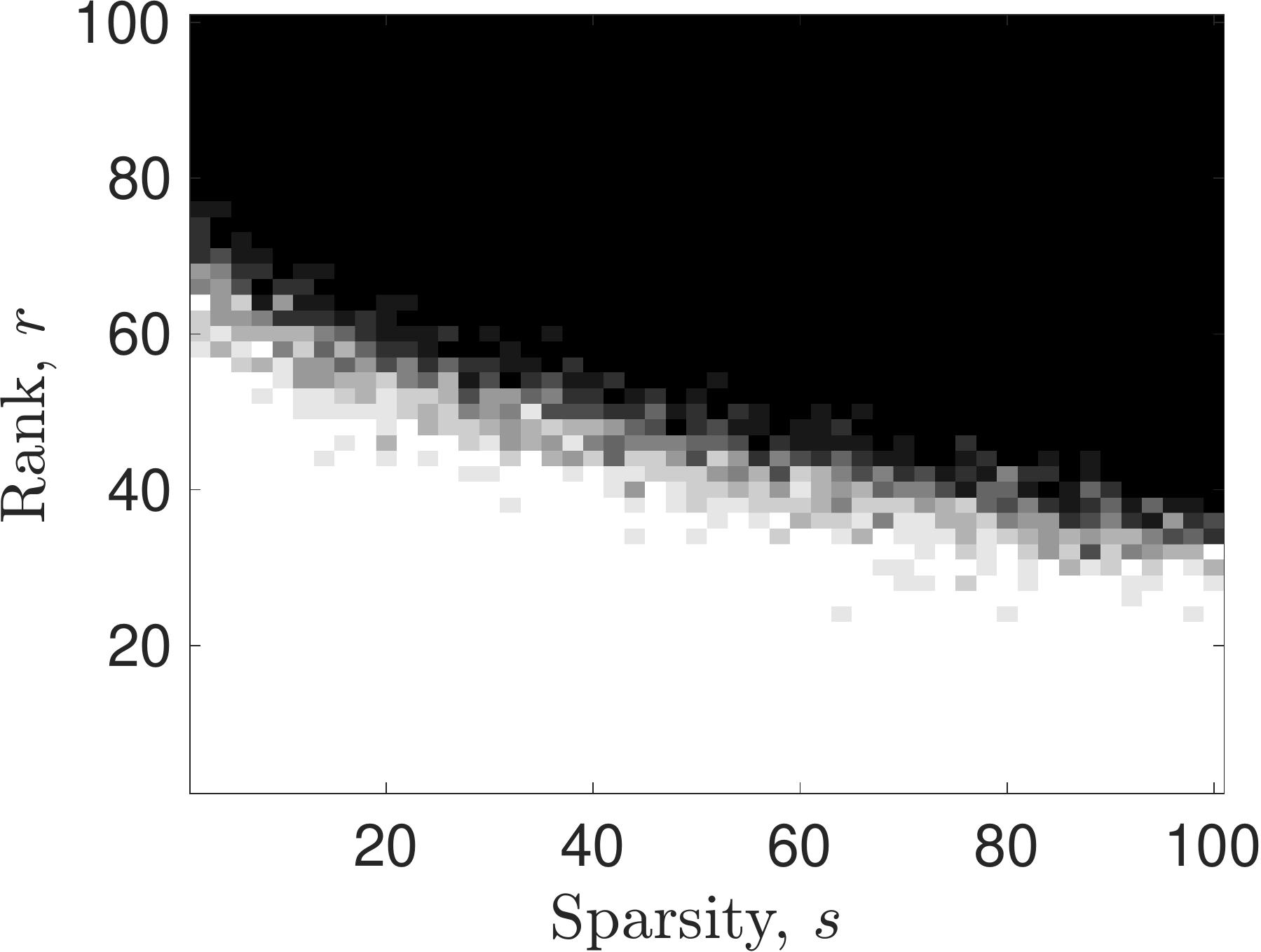} & \hspace{-0.4cm} \includegraphics[width=0.225\textwidth]{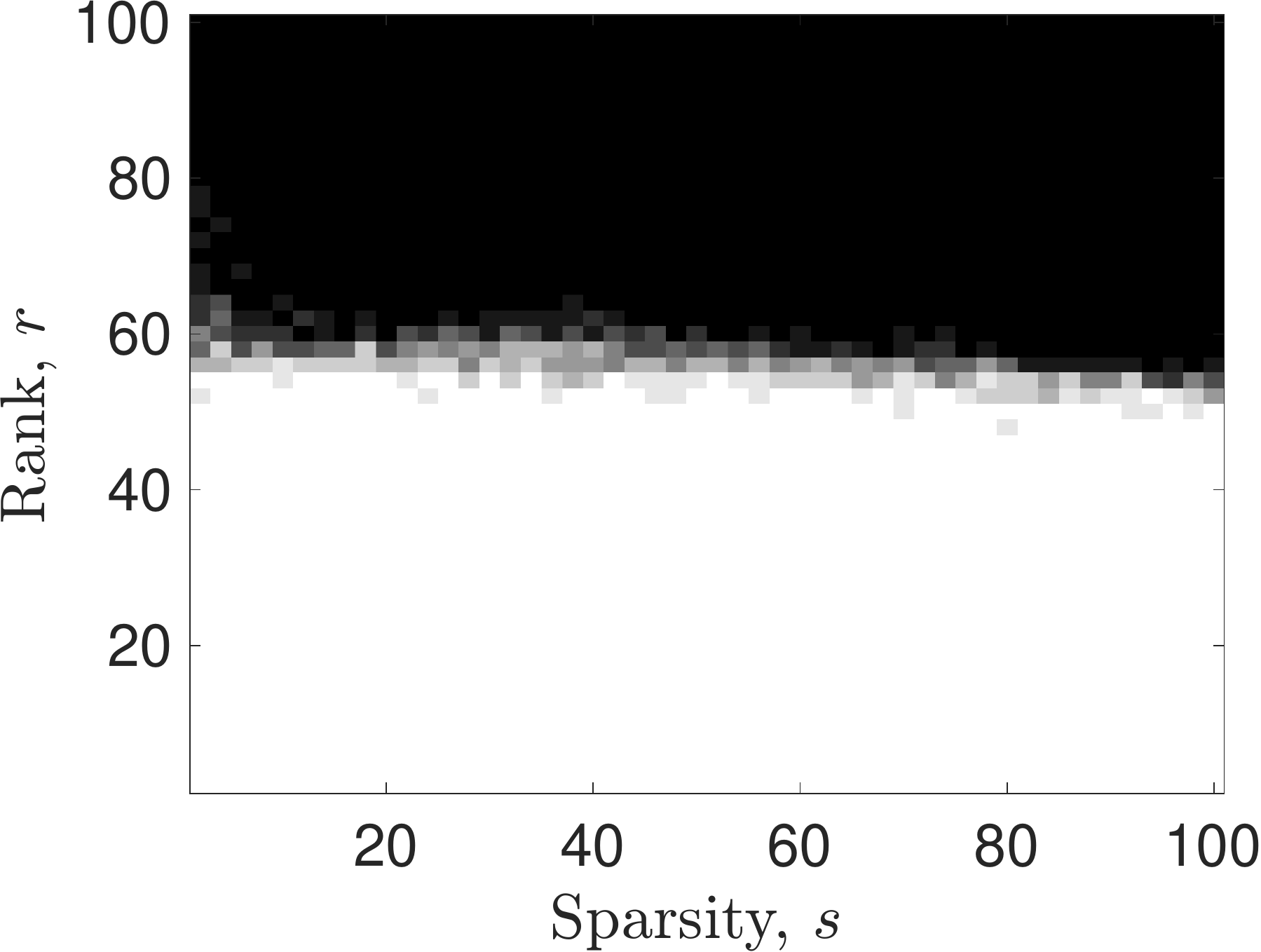} \vspace{-4pt}\\
		      ~~~~~~~~(c) & ~~(d) \vspace{-1pt}\\
		    \hspace{-0.2cm}{\small\rotatebox{90}{Recovery of $\mb{A}$ and $\mb{X}$}}
		    \includegraphics[width=0.225\textwidth]{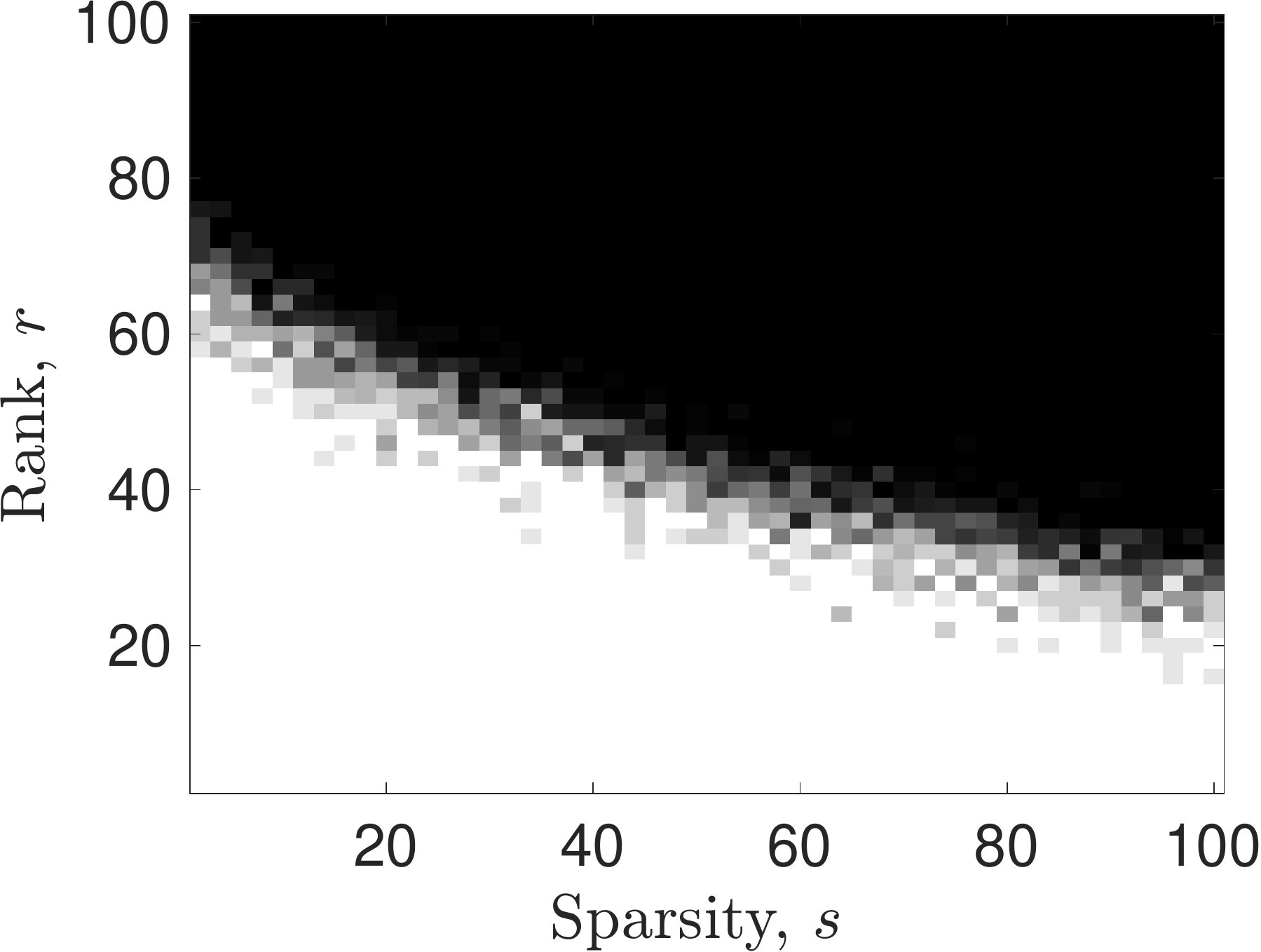} & \hspace{-0.4cm} \includegraphics[width=0.225\textwidth]{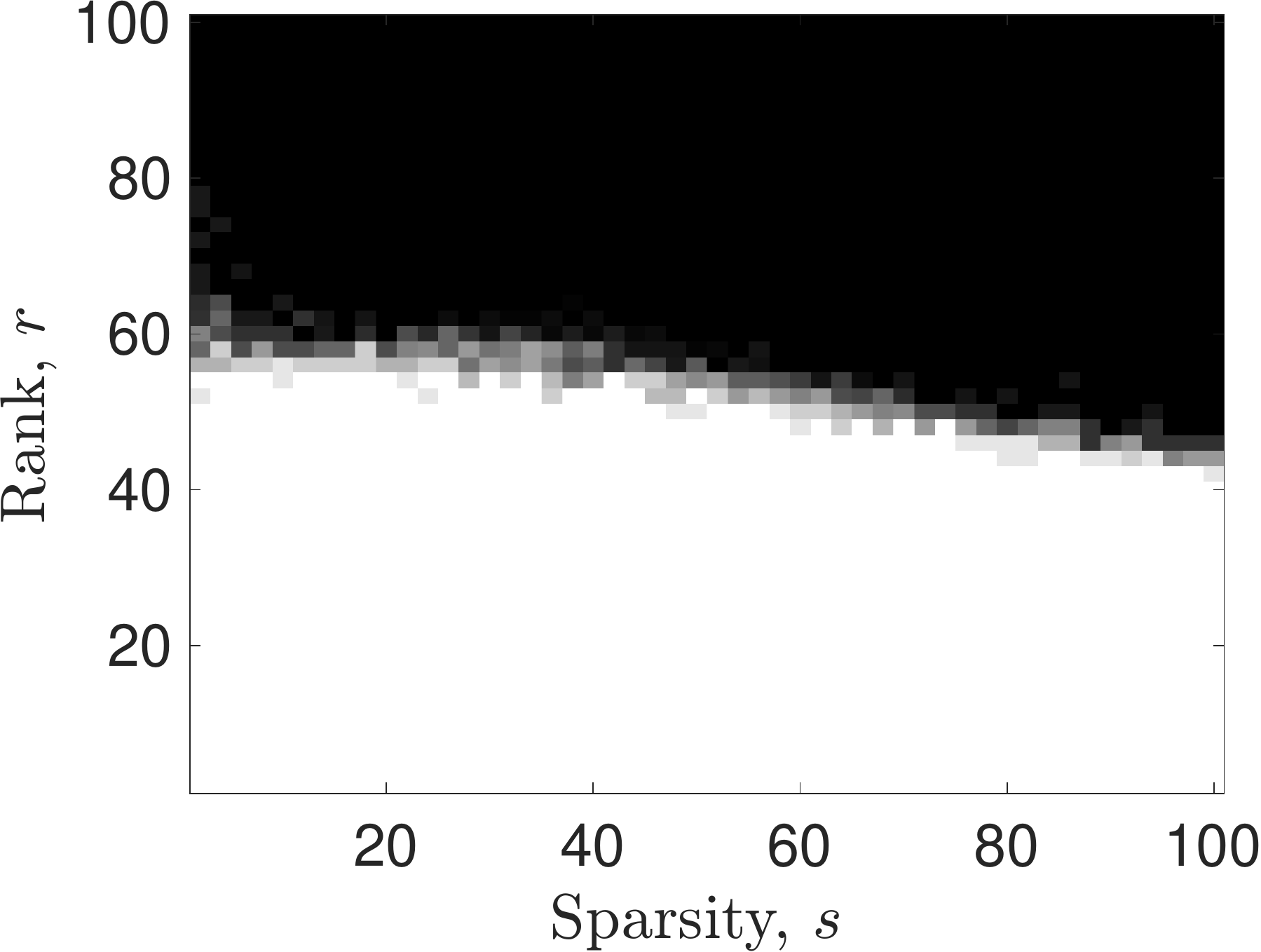} \vspace{-4pt}\\
		      ~~~~~~~~(e) & ~~~(f) 
		    \vspace{-0.4cm}
		     \end{tabular}}
		     \caption{Recovery for varying rank of {\footnotesize$\mb{X}$}, sparsity of {\footnotesize $\mb{A}$} and number of dictionary elements in {\footnotesize $\mb{R}$} as per Thm.~\ref{theorem}. Each plot shows average recovery across 10 trials for varying ranks (y-axis) and sparsity (x-axis) up to {\footnotesize $s \leq m$}, white region representing correct recovery, for $n = m = 100$. We decide success if {\footnotesize $\| \mb{X} - \hat{\mb{X}}\|_F/\| \mb{X} \|_F \leq 0.02$} and {\footnotesize $\| \mb{A} - \hat{\mb{A}}\|_F/\| \mb{A} \|_F \leq 0.02$}, where {\footnotesize $\hat{\mb{X}}$} and {\footnotesize $\hat{\mb{A}}$} are the recovered {\footnotesize $\mb{X}$} and {\footnotesize $\mb{A}$}, respectively. Panels (a)-(b) show the recovery of the low-rank part {\footnotesize $\mb{X}$} and (c)-(d) show the recovery of the sparse part with varying dictionary sizes $d = 5 ~\text{and}~ 150$, respectively. Panels (e) -(f) show the region of overlap where both {\footnotesize $\mb{X}$} and {\footnotesize $\mb{A}$} are recovered successfully.  Also, the predicted trend between rank {\small $r$} and sparsity {\small $s$} as per Thm.~\ref{theorem}, eq.\eqref{rankSpar1} and eq.\eqref{rankSpar2} is shown in red in panels (a-b).}
		     \vspace{-0.71cm}
		     \label{fig:phaseTr}
	\end{figure}
	\stopcompact{footnotesize} 	 
 Our analysis in the previous section shows that depending upon the size of the dictionary {\small$\mb{R}$}, if the conditions of Thm.~\ref{theorem} are met, a convex program which solves eq.\eqref{optProb} will recover the components {\small $\mb{X}$} and {\small$\mb{A}$}. In this section, we empirically evaluate the claims of Thm.~\ref{theorem}. To this end, we employ the accelerated proximal gradient algorithm outlined in Algorithm~1 of \cite{Mardani2012} to analyze the phase transition in rank and sparsity for different sizes of the dictionary  {\small$\mb{R}$}. For our analysis, we consider the case where {\small$n = m = 100$}. Here, we generate the low-rank part  {\small$\mb{X}$} by outer product of two random matrices of sizes {\small$n \times r$} and {\small$m \times r$}, with entries drawn from the standard normal distribution. In addition, the non-zero entries ({\small $s$} in number), of the sparse component  {\small$\mb{A}$}, are drawn from the Rademacher distribution, also the dictionary  {\small$\mb{R}$} is drawn from the standard normal distribution, then its columns are normalized. Phase transition in rank and sparsity over {\small$10$} trials for dictionaries of sizes {\small$d = 5$ } (thin) and {\small$d = 150$} (fat), corresponding to our theoretical results, and for all admissible levels of sparsity are shown in Fig.~\ref{fig:phaseTr} and  Fig.~\ref{fig:phaseTr_full}, respectively. 
 
   \startcompact{footnotesize}
    \stopcompact{footnotesize}
    \vspace{10pt}
 Fig.~\ref{fig:phaseTr} shows the recovery of the low-rank part {\small$\mb{X}$} in panels (a-b), while panels (c-d) show the recovery for the sparse component {\small$\mb{A}$}, for {\small$d = 5~\normalsize\text{and}~ 150$}, respectively. Next, panels (e-f) show the region of overlap between the low-rank recovery and sparse recovery plots, for {\small $d = 5$} and {\small $d = 150$}, respectively. This corresponds to the region in which both {\small$\mb{X}$} and {\small$\mb{A}$} are recovered successfully. Further, the red line in panels (a) and (b) show the trend predicted by our analysis, i.e., eq.\eqref{rankSpar1} and eq.\eqref{rankSpar2}, \edit{with the parameters hand-tuned for best fit}. Indeed, the empirical relationship between rank and sparsity has the same trend as predicted by Thm.~\ref{theorem}.%

Similarly,  Fig.~\ref{fig:phaseTr_full} shows the recovery of the low-rank part {\small$\mb{X}$} in panels (a-b), while panels (c-d) show the recovery for the sparse component {\small$\mb{A}$} for {\small$d = 5~\normalsize\text{and}~ 150$}, respectively, for a much wider range of global sparsity {\small $s$}. Indeed, these phase transition plots show that we can successfully recover the components for sparsity levels much greater than those put forth by the theorem. This can be attributed to the deterministic analysis presented here. To this end, \edit{we conjecture} that a randomized analysis of the problem can potentially improve the results presented here.

 %
 %
\vspace{-12pt}
\section{Conclusions}
\label{Conclusion}
\vspace{-10pt}
We analyze a dictionary based generalization of Robust PCA. Specifically, we extend the theoretical guarantees presented in \cite{Mardani2012} to a setting wherein the dictionary  {\small$\mb{R}$} may have arbitrary number of columns, and the coefficient matrix  {\small$\mb{A}$} has \textit{global} sparsity of {\small $s$}, i.e.  {\small$\|\mb{A}\|_0 = s \leq s_{\text{max}}$}. We generalize the results by assuming {\small$\mb{R}$} to be a \textit{frame} for the \textit{thin} case, to obey the RIP condition for the \textit{fat} case, and eliminate the orthogonality constraints on the rows of the dictionary {\small$\mb{R}$} and the sparsity constraint on the rows of the coefficient matrix {\small$\mb{A}$} (as in \cite{Mardani2012}), rendering the results useful for a potentially wide range of applications. Further, we provide empirical evaluations via phase transition plots in rank and sparsity corresponding to our theoretical results. We draw motivations from the promising phase transition plots, beyond the sparsity level tolerated by our analysis, and propose randomized analysis of the problem to improve the upper-bound on the sparsity as a future work. 

	\bibliographystyle{IEEEbib}
	\bibliography{referLR}

\end{document}

%% file: commands.tex
\newcommand{ \mb}[1]{\mathbf{#1}}

\newcommand{ \mc}[1]{\mathcal{#1}}